\documentclass[runningheads]{llncs}
\usepackage{graphicx}
\usepackage{cite}
\usepackage{multirow}
\usepackage{amsmath}
\usepackage{rotating}
\usepackage{amsfonts}
\usepackage{hyperref}
\usepackage{dirtytalk}

\begin{document}
\title{Conformance Checking for a Medical Training Process Using Petri net Simulation and Sequence Alignment}
%
%
\author{An Nguyen\thanks{Corresponding author}
\and 
Wenyu Zhang\thanks{Equal contribution with An Nguyen}
\and
Leo Schwinn
\and
 Bjoern Eskoﬁer
 }
\authorrunning{A. Nguyen, W. Zhang et al.}
%

\titlerunning{Conformance Checking for a Medical Process via Sequence Alignment} 

\institute{Department of Computer Science, Friedrich-Alexander-University Erlangen-Nürnberg (FAU), Erlangen, Germany 
\email{\{an.nguyen, wenyu.zhang, leo.schwinn, bjoern.eskofier\} @fau.de}\\
}

\maketitle              
\begin{abstract}
Process Mining has recently gained popularity in healthcare due to its potential to provide a transparent, objective and data-based view on processes.
Conformance checking is a sub-discipline of process mining that has the potential to answer how the actual process executions deviate from existing guidelines. 
In this work, we analyze a medical training process for a surgical procedure. Ten students were trained to install a Central Venous Catheters (CVC) with ultrasound. Event log data was collected directly after instruction by the supervisors during a first test run and additionally after a subsequent individual training phase.
In order to provide objective performance measures, we formulate an optimal, global sequence alignment problem inspired by approaches in bioinformatics. Therefore, we use the Petri net model representation of the medical process guideline to simulate a representative set of guideline conform sequences. Next, we calculate the optimal, global sequence alignment of the recorded and simulated event logs. Finally, the output measures and visualization of aligned sequences are provided for objective feedback.

\end{abstract}


\section{Introduction}
Process mining deals with the extraction of insights from event logs. Its major groups of methods are process discovery, conformance checking and process enhancement \cite{pm_book}. Event logs analyzed in process mining are usually grouped into so-called cases or traces, which consist of a sequence of activities and timestamps.
\par
Process mining recently gained popularity for healthcare applications, as indicated in the reviews \cite{review_1} and \cite{review_2}. There exist many guidelines for the vast amount of processes in healthcare. A common question is whether the process executions are conformable with existing guidelines. 
\par 
To very this, conformance checking could be applied. Conformance checking is an approach that aims to check how the recorded event log data from actual process executions \say{match} a given normative model which represents guidelines. Such models are often given in workﬂow languages like Petri nets and BPMN \cite{pm_book}.
\par
Lira et al. \cite{ref_article3} introduced the idea to provide process-oriented feedback for a procedural training. They created feedback reports based on the commercial process mining software Celonis \footnote{\url{http://www.celonis.com}} and its implemented process discovery methods. The discovered models were used to identify rework, undesired order of activities, and performance issues in terms of duration.
\par 
Bose et al.\cite{ref_article1,ref_article2} applied trace alignment methods inspired by bioinformatics to analyze event log data. Rental agency and telephone repair logs were used as examples. They demonstrated that the process mining discipline could beneﬁt from alignment techniques. Their goal was to find a global alignment between all cases within one event log by applying multiple alignment.   
\par
Here, we propose an alternative analysis of a medical training process for the placement of Central Venous Catheter (CVC) with ultrasound compared to \cite{ref_article3}. Event log data of student trials and a Petri net which represents the process guidelines were provided by the 2019 Conformance Checking Challenge (CCC19) \cite{CCC19}. We aim to provide objective performance measures for both students and instructors by formulating the conformance checking problem as an optimal global alignment problem. Therefore, we create a normative event log by simulating a given Petri net, which represents the medical process guideline. Next, we compute the maximum global alignment of each case in the student event log with all cases in the simulated event log based on the maximum identity. The resulting alignment results are concisely summarized with visualization techniques inspired by the bioinformatics community to evaluate the student performance. The proposed framework is easy to apply and provides objective and interpretable measures of the student performance.

\section{Preliminaries}
\label{sec:prelim}
In the following we give some preliminary definitions
which we will refer to in the remainder of this work. 

\begin{definition}{\textbf{(event logs and sequences)}}
\label{def1}
Let $\mathbb{A}$ be the set of all $activities$ of a process $\mathbb{P}$. A sequence of activities is described by a mapping $\sigma \in \{1, ..., n\}   \to \mathbb{A}$, where $n$ is the length of the sequence ($len(\sigma) = n$). $\sigma(i)$ denotes the $i^{th}$ element of the sequence, for $1 \leq i \leq n$. A sequence $\sigma$ can be denoted by $\left\langle a_1, a_2,  ..., a_n\right\rangle$, where $a_i = \sigma(i)$. An event log $\mathbb{L}$ is a multiset of sequences $\sigma$. We denote $\sigma_{i, \mathbb{L}}$ as the $i^{th}$ sequence in the event log $\mathbb{L}$. $N_{seq} = |\mathbb{L}|$ and $N_{act} = |\mathbb{A}|$ are the number of sequences and unique activities in an event log $\mathbb{L}$ and process $\mathbb{P}$ respectively.
\end{definition}

\noindent We interchangeably refer to one student execution as a case, trace or sequence. 

\begin{definition}{\textbf{(stages and stage event log)}}
\label{def2}
Let $\mathbb{A}$ be the set of all $activities$ of a process $\mathbb{P}$. We map each $activity$ in $\mathbb{A}$ to a finite set of stages $S_j \in \mathbb{S}$ for $1 \leq j \leq m$, where $m$ denotes the number of stages in a process $\mathbb{P}$. Assume an event log $\mathbb{L}$ over a set of $activities$ $\mathbb{A}$ with stages $S_j \in \mathbb{S}$. We define the 'Stage event log'~$\mathbb{L}_{S_j}$ as the event log, where the $sequences$ $\sigma_{i, \mathbb{L}_{S_j}} = \left\langle a_{S_j, first}, ..., a_{S_j, last}\right\rangle$ are derived from $sequences$  $\sigma_{i, \mathbb{L}} = \left\langle a_{1}, ..., a_{n}\right\rangle$ with $a_{S_j, first}$ and $a_{S_j, last}$ being the first and last $activity$ in $\sigma_{i, \mathbb{L}} $ assigned to stage $S_j$ respectively. $\sigma_{i, \mathbb{L}_{S_j}}$ includes all activities from $\sigma_{i, \mathbb{L}}$ including and between $a_{S_j, first}$ and $a_{S_j, last}$.
\end{definition}

\begin{definition}{\textbf{(alignment)}}\label{def3}
Given two sequences $\sigma_1 = \left\langle a_1, a_2, ..., a_n \right\rangle\ $ and \mbox{ $\sigma_2 = \left\langle b_1, b_2, ..., b_m\right\rangle\ $} of lengths n and m respectively. An alignment is the assignment of gaps denoted as '-' to positions 0, .., max(m, n) in $\sigma_1$ and $\sigma_2$, such that each activity in one sequence is lined up with either an activity or gap in the other sequence. The resulting sequences based on such alignment are referred to as $\sigma_1^*$ and $\sigma_2^*$.
\end{definition}

\begin{definition}{\textbf{(identity)}}\label{def5}
Given two aligned sequences $\sigma_1^*$ and $\sigma_2^*$ of length N. We define the identity as the percentage of matching activities in these sequences.

$$identity  = \frac{100}{N} \cdot \sum{[\sigma_1^*(i) = \sigma_2^*(i)}] , ~\sigma_1^*(i)  ~ and  ~ \sigma_2^*(i) \in A \eqno(2)$$

\end{definition}

\section{Data}
\label{sec:data}
The data analyzed in this study was provided by the Conformance Checking Challenge 2019 (CCC19) \cite{CCC19}. Ten students were trained to install a Central Venous Catheter (CVC) with ultrasound. Event log data was both collected at pre- and post-trial via tagged video recordings \cite{ref_article3}. The process can be classified into six subsequent stages:' Operator and Patient Preparation' ($S1$),' Ultrasound Preparation' ($S2$),' Locate Structures' ($S3$),' Venous puncture' ($S4$),' Install Guidewire' ($S5$), and' Install Catheter' ($S6$). Each of these stages includes a set of activities which are displayed in Table \ref{table:activities}, where every activity is represented by a specified letter. Activities are color-coded based on their corresponding stages as depicted in Table \ref{table:activities}.
\par
Students and instructors are particularly interested in the comparison of performance between both trials with respect to the process guideline. The normative process model represented as a Petri net, is shown in Figure  \ref{fig:process}, including all activities and stages of the process. Note that the activities in the Petri net model 'INVISIBLE No good position' and' INVISIBLE No Return' were not recorded in the given event log.

\begin{table}[!h]
\centering
\resizebox{0.65\columnwidth}{!}{%
\begin{tabular}{c|c|c}
      \hline
      Stage (Color Encoding) & Activity & Abbreviations  \\
      \hline
      \multirow{5}{*}{S1 (Purple)} & Prepare Implements & a \\
      & Hand washing & b \\
      & Get in sterile clothes & c \\
     & Clean puncture area  & d \\
      & Drap puncture area & e \\
        \hline
      \multirow{5}{*}{S2 (Blue)} & Ultrasound conﬁguration  & f \\
      & Gel in probe  & g \\
     & Cover probe & h \\
      & Put sterile gel  & i  \\
      & Position Probe & j \\
      \hline
\multirow{4}{*}{S3 (Green)} & Position patient  & k \\
      & Anatomic identiﬁcation   & l\\
     & Doppler identiﬁcation &  m \\
      & Compression identiﬁcation   & n  \\
      \hline
      \multirow{3}{*}{S4 (Orange)} & Anesthetize & o \\
      & Puncture  & p \\
     & Blood return & q \\
      \hline
      \multirow{7}{*}{S5 (Red)} & Drop probe  & r  \\
     & Remove syringe  & s \\
      & Guidewire install & t \\
      & Remove trocar & u \\
      & Check wire in short axis & v \\
      & Check wire in long axis & w\\
      & Wire in good position  & x \\
      \hline
  \multirow{5}{*}{S6 (Yellow)} & Widen pathway  & y \\
      & Advance catheter  & z\\
      & Remove guidewire  & 0 \\
     & Check ﬂow and reﬂow   &  1 \\
      & Check catheter position & 2  \\
        \hline
\end{tabular}%
}
\caption{\label{table:activities}All activities of the process with the assigned stages and abbreviations.}
\end{table}

\section{Methods}
\label{sec:meth}
In this section, we give an overview of the methods used in this work.

\subsection{Sequence Alignment Framework}
\begin{figure*}[tbh]
\centering
\includegraphics[width=1\textwidth]{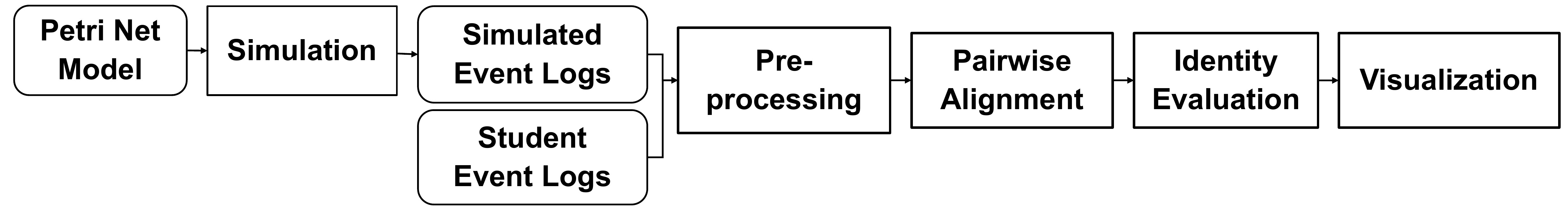}
\caption{Proposed framework for sequence alignment and visualization.}
\label{fig:framework}
\end{figure*}

Figure \ref{fig:framework} presents the overall framework. The optimal alignment between each individual case of the event log recorded from the process executions of the students $\mathbb{L}_{student}$ and all simulated normative cases in $\mathbb{L}_{norm}$ was computed to provide objective performance measures. Inspired by the bioinformatics viewer \say{AliView} \cite{aliview}, aligned process sequences are represented in an intuitive format. In order to give more detailed feedback based on the performance in the different stages ($S1$, ..., $S6$), we extracted the corresponding event logs $\mathbb{L}_{student_{S_{i}}}$ and $\mathbb{L}_{norm_{S_{i}}}$ for $1 \leq i \leq 6$ according to Deﬁnition \ref{def2}. Finally, the aligned sequences and identity for the whole process and each of the stages are visualized as feedback for the students and instructors. The source code is available on GitHub\footnote{\url{https://github.com/annguy/sequence-alignment-conformance-checking}}.

\subsection{Event Log Simulation}
We used the Petri net, as shown in Figure \ref{fig:process}, in combination with the ProM \cite{ProM} plug-in \say{Perform a simple simulation of a stochastic Petri net} to generate a representative event log $\mathbb{L}_{norm}$ for the medical process guidelines at hand. The shortest and longest student sequence in $\mathbb{L}_{student}$ were of lengths $26$ and $59$ respectively. Therefore, we set the maximum number of simulated activities as $65$. And the number of simulated sequences was set as $1.000$ to cover most case variants. We evaluated how many sequences should be simulated for $\mathbb{L}_{norm}$ in order to get a good representation of the guidelines. Therefore, we tried $10.000$ and $100.000$ simulations which all yielded the same final alignment results as for $1.000$. This implies that $1.000$ simulations are sufficient for this process.
\par
In a pre-processing step, only simulated sequences that completed the process (ended with the activity' Check catheter position') were kept for further analysis in $\mathbb{L}_{norm}$. Furthermore, duplicated sequences and the activities 'INVISIBLE No good position', and' INVISIBLE No Return' were removed from $\mathbb{L}_{norm}$. After pre-processing, $315$ different simulated sequences remained in $\mathbb{L}_{norm}$. 

\subsection{Sequence Alignment}
The \textbf{Needleman-Wunsch algorithm} \cite{needle} computes an optimal global alignment such that both sequences are aligned from their first position through their last residue, given two sequences $\sigma_1 = \left\langle a_1, a_2, ..., a_n \right\rangle\ $ and \mbox{ $\sigma_2 = \left\langle b_1, b_2, ..., b_m\right\rangle\ $}. The algorithm essentially divides the full sequences into a series of sub-sequences. For each alignment, a dynamic programming matrix $F$ and a traceback matrix $T$ are generated. The entry in row $i$ and column $j$ is denoted here by $F_{(i,j)}$ ($i = 0,...,n, ~ j=0,...,m$) and $T_{(i,j)}$ ($i = 0,...,n, ~ j=0,...,m$). First, we start with a zero in the first row and first column $F_{(0,0)}$. Next, we compute the rest of elements in $F_{(i,j)}$ as the maximum of three possible values as follow,
\[
F_{(i,j)}=max
\begin{cases}
F_{(i-1,j-1)} + s_{(i,j)}\\
F_{(i-1,j)} - d\\
F_{(i,j-1)} - d 
\end{cases} \eqno(1)
\]

\begin{figure}[!h]
\begin{center}
\includegraphics[scale=0.4]{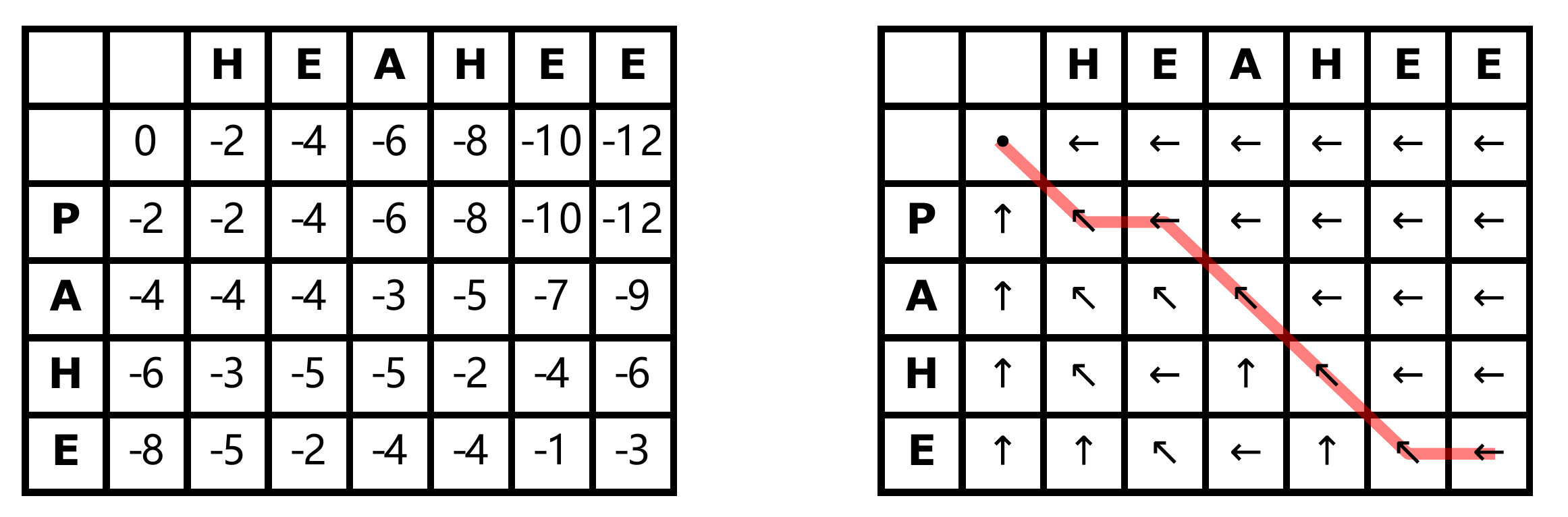}
\caption{Application of the Needleman-Wunsch algorithm for sequences $\sigma_1$ = "HEAHEE" and $\sigma_2$ = "PAHE" to compute the dynamic programming matrix $F$ (left) and traceback matrix $T$ (right). The red line represents transcribing path. The aligned sequences are $\sigma_1^*$ = "HEAHEE" and $\sigma_2^*$ = "P-AHE-". The resulting identity is $50\%$.}
\label{global_matrices}
\end{center}
\end{figure}

$s_{(i,j)}$ refers to the similarity of characters $a_i$ and $b_j$ in corresponding sequence. If $a_i = b_j$, $s_{(i,j)} = match ~ reward$, and if $a_i \neq b_j$, $s_{(i,j)} = mismatch ~ penalty$. $d$ represents $gap ~ penalty$. The final score of a pairwise alignment can be defined as the sum of the scores of the edit operations across all sub-sequences in the alignment. Optimal alignment can be considered to be the one with the maximum score. Inserting arrows in $T$ is according to the process to decide which cell each score was derived from the $F$ matrix. For example, if deriving the score for a given cell in $F$ from the cell diagonally up and to the left, a diagonal arrow is added in $T$. Figure \ref{global_matrices} shows an example of globally aligning two sequences. 

Transcribing starts at the bottom-right in $T_{(n,m)}$ to the cell on the top left in $T_{(0,0)}$ by following the direction of the arrows. We consume a pair of characters onto the front of the alignment: elements $a_i$ and $b_j$ from $\sigma_1$ and $\sigma_2$ if we encounter a diagonal arrow, $a_i$ and gap symbol '-' if we encounter a horizontal arrow, '-' and $b_j$ if we encounter a vertical arrow. The traceback procedure depicted as red line in Figure \ref{global_matrices} will retrieve only one of the alignments that gives the best score.

In this work we used a Python implementation \footnote{https://gist.github.com/aziele/6192a38862ce569fe1b9cbe377339fb} of the algorithm. We applied the Needleman-Wunsch algortihm with the scores: $match = 1$, $gap = - 2$ and $mismatch = -2$. For each individual sequence in the student event logs $\mathbb{L}_{student}$ and $\mathbb{L}_{student_{S_i}}$, we calculated the optimal alignment with all sequences in the corresponding normative event logs $\mathbb{L}_{norm}$ and $\mathbb{L}_{norm_{S_i}}$ respectively. Specifically, we found the maximum identity (Definition \ref{def5}) of each sequence in $\mathbb{L}_{student}$ or $\mathbb{L}_{student_{S_i}}$ with all sequences in $\mathbb{L}_{norm}$ or $\mathbb{L}_{norm_{S_i}}$ respectively.

\subsection{Alignment Visualization}
\begin{figure*}[tbh]
\centering
\includegraphics[width=1\textwidth]{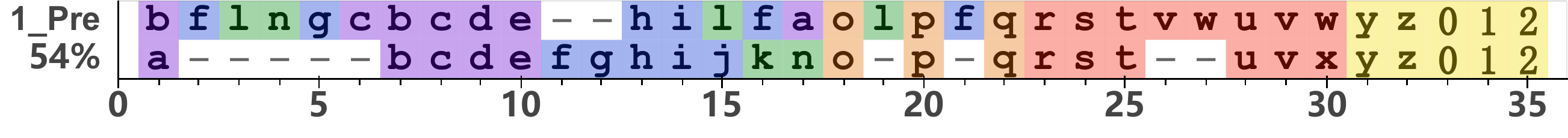}
\caption{A visualization example of pairwise alignment. The index or case ID and alignment information such as maximum identity are depicted on the left. The aligned sequences are plotted on the right. Colors are assigned to each stage as described in Table \ref{table:activities}.}
\label{test}
\end{figure*}

We provide an intuitive visualization of the resulting alignments using Bokeh \cite{Bokeh}. Similar to biological sequence alignment visualizations, our implementation can plot the aligned sequences. Vital components, deviations, and exceptions are quite evident in the resulting visualization. Figure \ref{test} exemplifies the visualization for the alignment of a specific student trace and simulated sequence with maximum identity. 

\section{Results}
\label{sec:res}

Based on the proposed framework, we present the results of sequence alignment and visualization in the subsequent sections.

\begin{figure*}[tbh]
\centering
\includegraphics[width=1\textwidth]{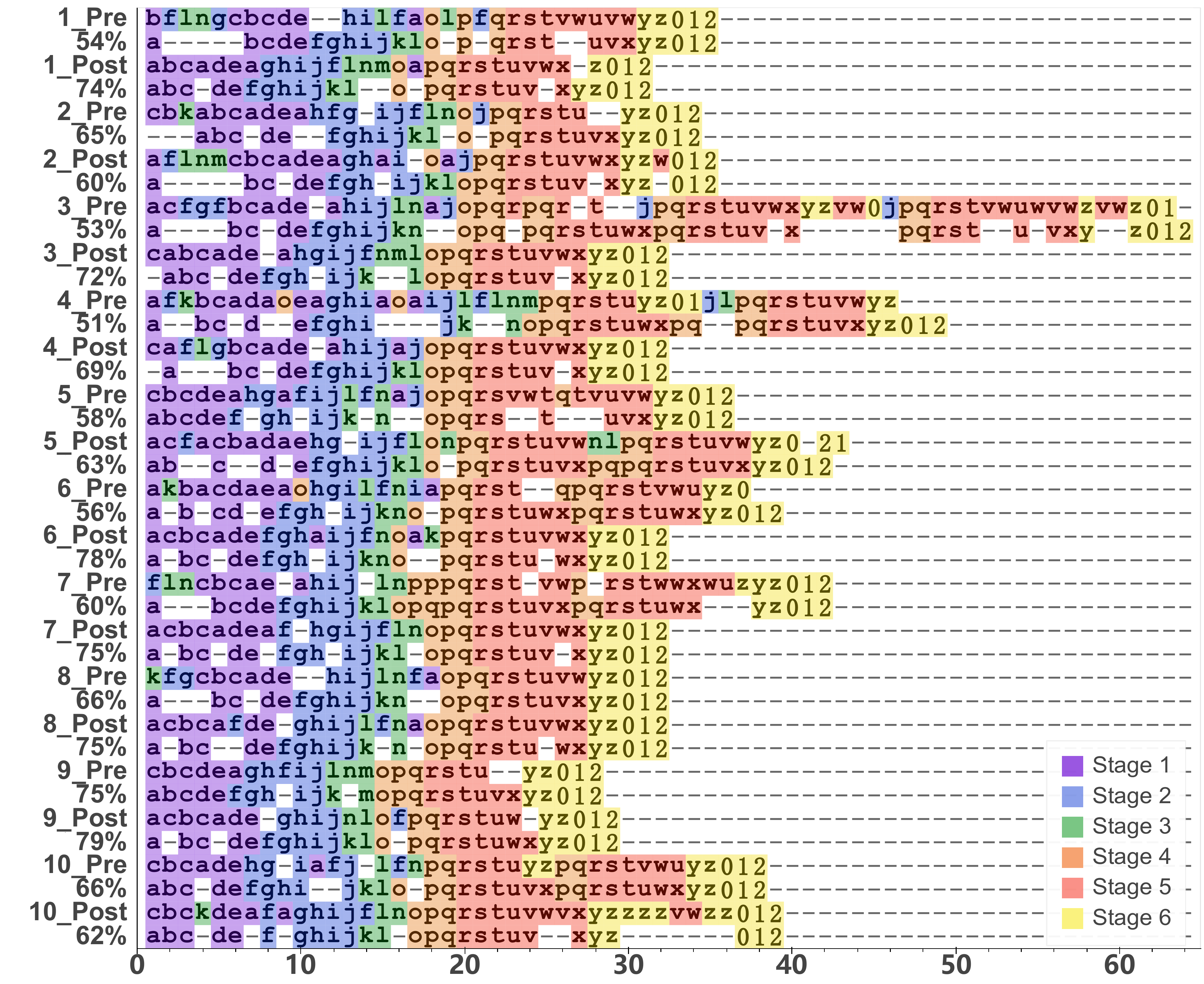}
\caption{The output of the Needleman-Wunsch algorithm for sequences of pre- and post-trials from $\mathbb{L}_{student}$ and sequences from $\mathbb{L}_{norm}$ with maximum identity. We set the parameters as $match = 1$, $gap = - 2$ and $mismatch = -2$. The label on the left for every row includes information about the case ID of every trial and maximum identity with aligned sequences from $\mathbb{L}_{norm}$.}
\label{needle_pairwise_alignments}
\end{figure*}

\begin{figure*}[tbh]
\centering
\includegraphics[width=1\textwidth]{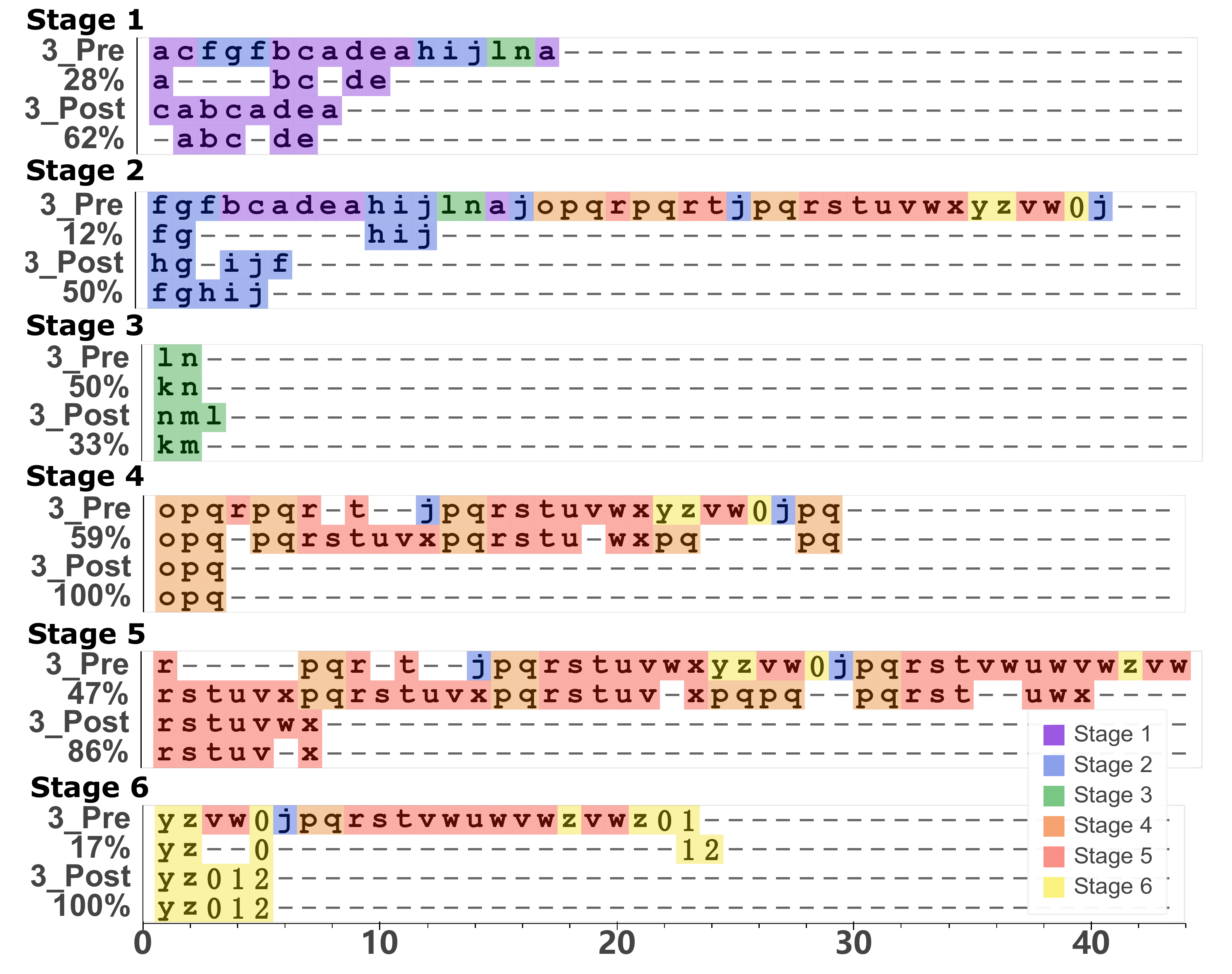}
\caption[]{The pairwise alignment of all 6 process stages for \say{3\_Pre} and \say{3\_Post}. Six subplots demonstrate the alignment of stage event logs $\mathbb{L}_{student_{S_i}}$ and $\mathbb{L}_{norm_{S_i}}$ ($1 \leq i \leq 6$) from top to bottom. Activities are color-coded according to their assigned stages.}
\label{student3_all_stages}
\end{figure*}

\subsection{Optimal Alignment Sequences}
 The optimal alignment for each students' pre- and post-trial for the whole process in $\mathbb{L}_{student}$ and sequences in $\mathbb{L}_{norm}$ are illustrated in Fig. \ref{needle_pairwise_alignments}. The maximum identity for each pairwise alignment of each student trial in $\mathbb{L}_{student}$ with $\mathbb{L}_{norm}$ are shown on the left. Deviations, exceptions, and recurring patterns are captured in the visualizations of the alignments. For example for student \say{3\_Pre}, it can be observed that the activities $fgf$ belong to $S2$ before the accomplishment of $S1$. $pq$ and $rstuvwx$ are recurring patterns in pre-trial of student $3$ and the aligned simulated sequence. Contrary, for \say{3\_Post}, no exceptional behavior and loops are visible, which indicates a significant improvement after the training period of student $3$. This is also reflected in the improved identity score ($72\%$ vs $53\%$).
 \par
We also provide individual alignment feedback of every process stage to identify progress and deficiency of every student. Figure \ref{student3_all_stages} exemplifies the pairwise alignment of student $3$ for pre- and post-trials on a stage level. We could observe the improvement in all stages except for $S3$. The activities' executions are almost all correct after the individual training period for $S4$, $S5$, and $S6$. The mismatch and omission of activities suggest that additional training and instruction for stages $S1$, $S2$, and $S3$ might be beneficial for student $3$. Similar feedback can easily be provided for other students by inspecting the corresponding visualization.

\subsection{Identity Analysis}
\begin{figure}[b!]
\begin{center}
\includegraphics[scale = 0.35]{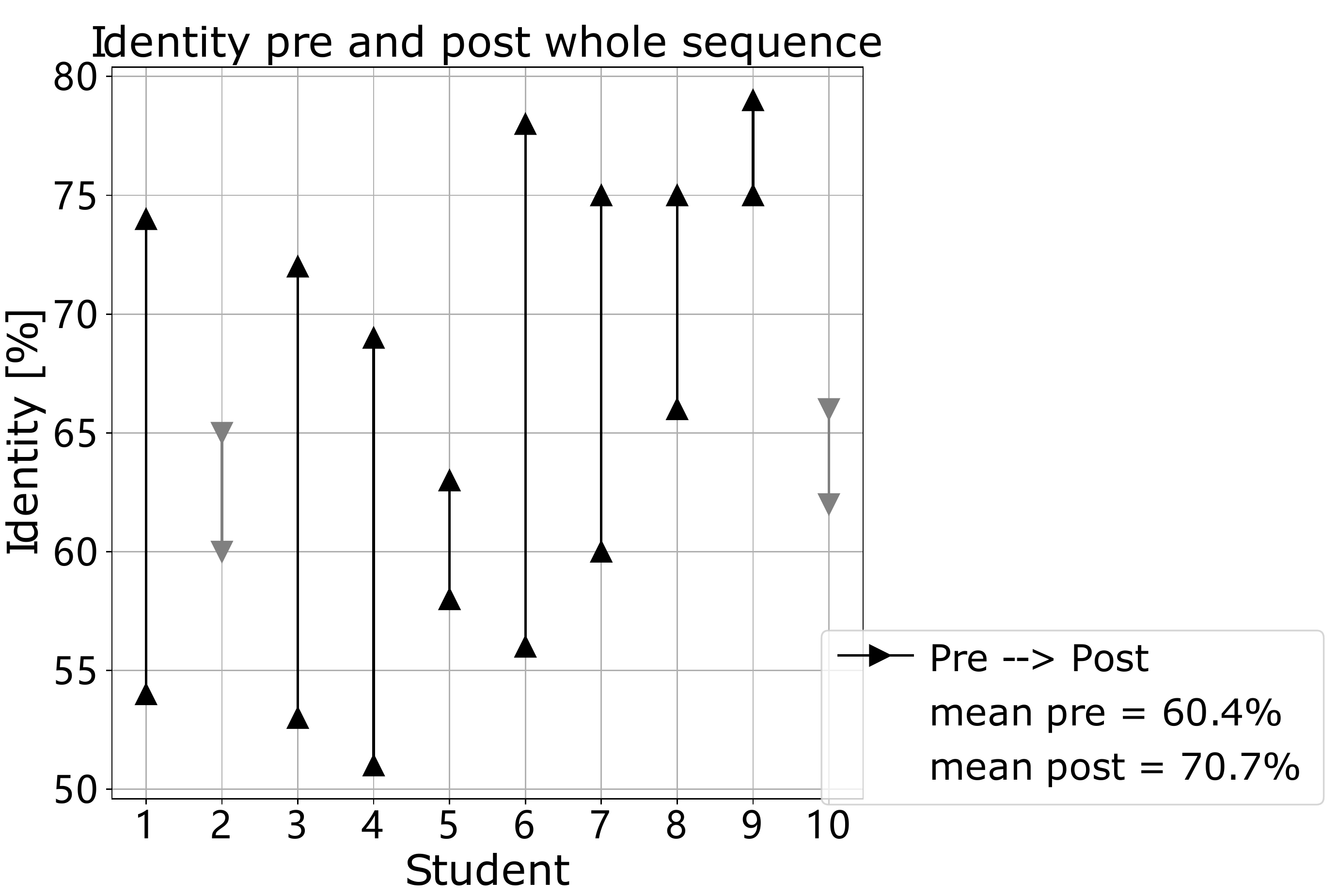}
\caption{Identity of the whole process for each students' pre- and post-trial. The markers are pointing from pre to post. Black indicates an improvement and gray a decline in performance from pre- to post-trial.} \label{fig_id_all}
\end{center}
\end{figure}

\begin{figure}[hb]
\includegraphics[width=1\textwidth]{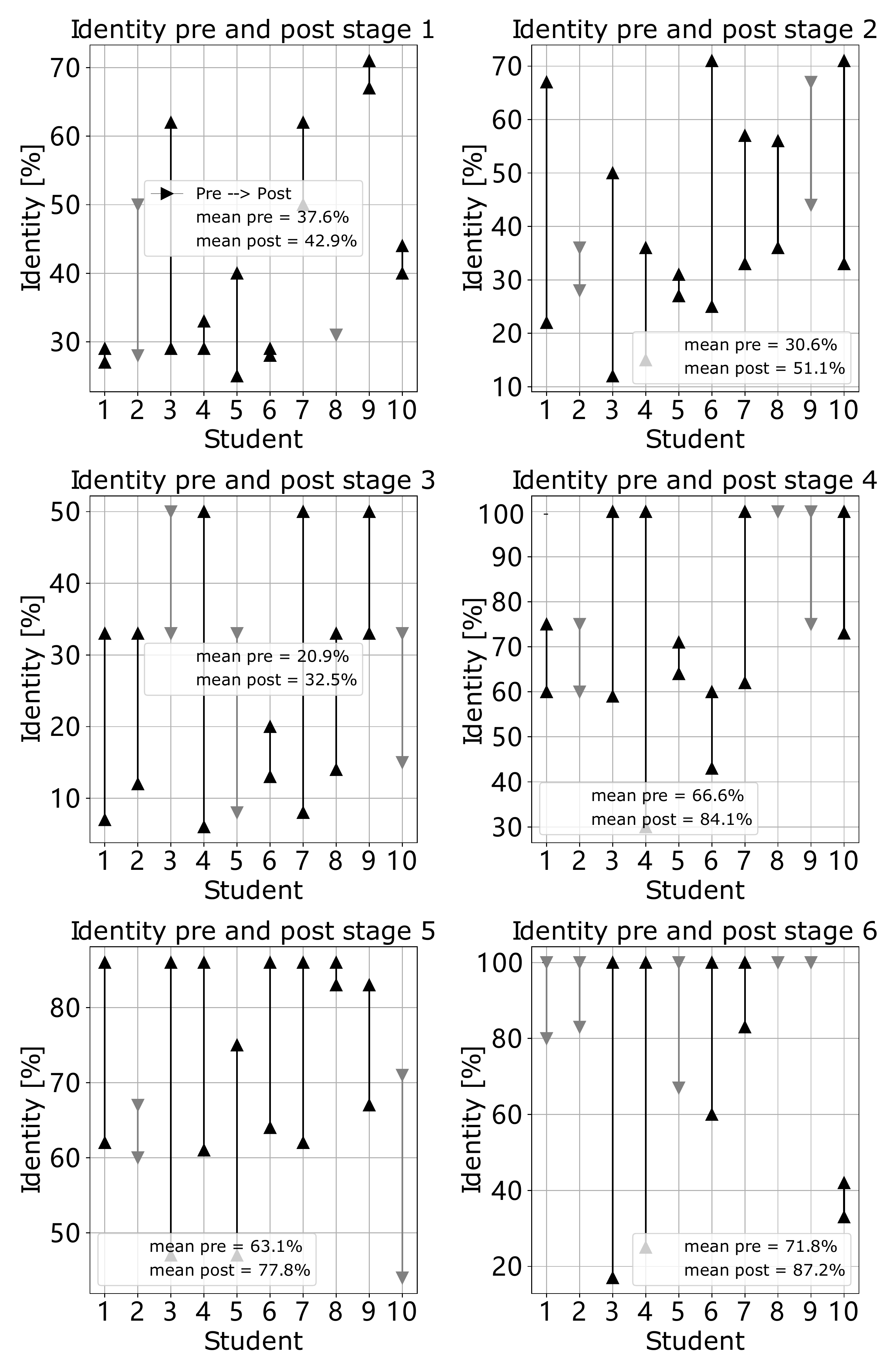}
\caption{Identity of all 6 process stages for each students' pre- and post-trial. The markers are pointing from pre to post. Black indicates an improvement and gray a decline in performance from pre-trial- to post-trial.} \label{fig_id_stages}
\end{figure}

Figures \ref{fig_id_all} and \ref{fig_id_stages} show the maximum identity of each sequence in $\mathbb{L}_{student}$ and $\mathbb{L}_{student_{S_i}}$ with  $\mathbb{L}_{norm}$ and $\mathbb{L}_{norm_{S_i}}$ respectively. For the whole process we observed that most students improved their performance in terms of measured conformance (identity) with the guidelines from pre-trial to post-trial. The mean overall improvement for the whole process is around $10\%$ (Fig. \ref{fig_id_all}). Additionally, an improvement for all $6$ stages in terms of computed identity can be seen in Figure \ref{fig_id_stages}. Considering the post-trial, students seem to perform best in stage $S_6$ \mbox{(mean identity = $87.2\%$)} and worst in stage $S_3$ (mean identity = $32.5\%$). Hence, a valid conclusion would be to intensify training stage $S_3$ in the CVC process.  

\section{Conclusion}
\label{sec:con}

\begin{sidewaysfigure}[htb!]
\includegraphics[width=\textwidth]{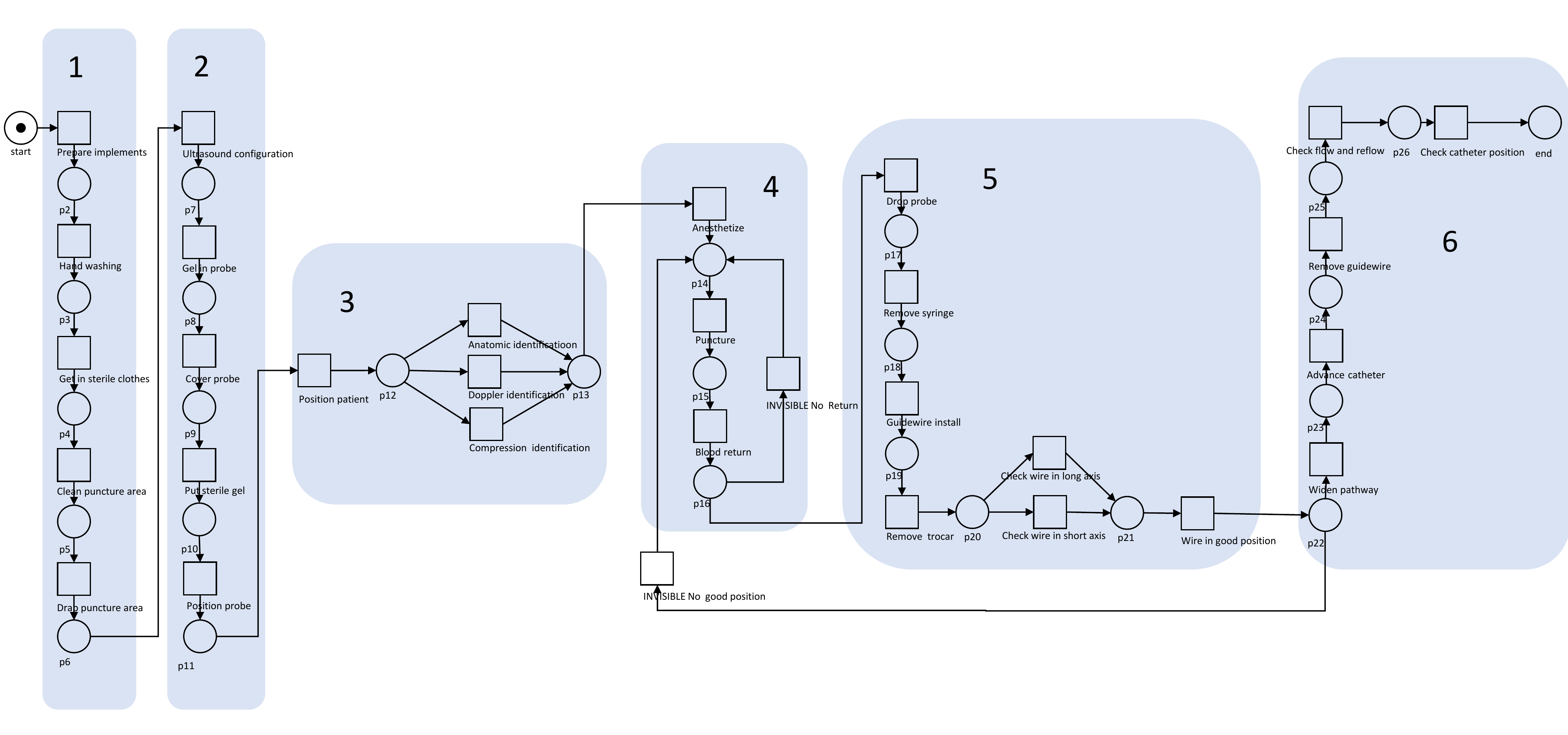}
\caption{Petri net representation of the normative model for the Central Venous Catheters (CVC) with ultrasound process. Activities are modeled as transitions and the initial marking includes one single token at the first place. The six subsequent stages of the process: 'Operator and Patient Preparation' (1) , 'Ultrasound Preparation' (2), 'Locate Structures' (3), 'Venous puncture' (4), 'Install Guidewire' (5) and 'Install Catheter' (6) are highlighted.} \label{fig:process}
\end{sidewaysfigure}

In this work, we proposed a framework to analyze the student performance for a medical training process - installing a Central Venous Catheters with ultrasound. A provided Petri net (Fig. \ref{fig:process}) modeled the guidelines for this process. We used this Petri net to simulate a normative event log for the overall process $\mathbb{L}_{norm}$ and its stages $\mathbb{L}_{norm_{S_i}}$. The maximum identity of each sequence of the student executions in ($\mathbb{L}_{student}$ and $\mathbb{L}_{student_{S_i}}$) with the normative event logs ($\mathbb{L}_{norm}$ and $\mathbb{L}_{norm_{S_i}}$) were computed via the Needleman-Wunsch algorithm. The resulting alignment results can help to provide an in depth feedback for both students and instructors. The resulting identity scores and visualizations are consistent with the expectations that the students improved after a longer training period. 
\par
Sequence alignment and visualization could yield potential advantages in the analysis of medical training processes by providing objective and interpretable feedback. The proposed framework can be applied easily to similar problems where an event log of process executions and a model which represents the guidelines are available. 
As a next step, it would be interesting to get feedback from students and instructors if they find the provided performance measures and visualizations helpful. Furthermore, a comparison with more traditional conformance checking approaches could be performed. 

\end{document}